\pdfoutput=1

\documentclass[11pt]{article}
\usepackage{float}

\usepackage[]{acl2023}

\usepackage[utf8]{inputenc}
\usepackage{pgfplots}
\usepackage{dsfont}
\DeclareUnicodeCharacter{2212}{−}
\usepgfplotslibrary{groupplots,dateplot}
\usetikzlibrary{patterns,shapes.arrows}
\pgfplotsset{compat=newest}

\usepackage{tikzscale}

\usepackage{multirow, colortbl}

\usepackage{tabularx,booktabs}
\usepackage{makecell}

\usepackage[normalem]{ulem}
\useunder{\uline}{\ul}{}

\usepackage{graphicx}

\definecolor{ablation6}{HTML}{fcefed}
\definecolor{ablation_tie}{HTML}{fce3e1}

\definecolor{ablation5}{HTML}{fcd8d4}
\definecolor{ablation4}{HTML}{FBC3BC}
\definecolor{ablation3}{HTML}{F7A399}
\definecolor{ablation2}{HTML}{F38375}
\definecolor{ablation1}{HTML}{EF6351}

\usepackage[]{algpseudocode}
\usepackage[]{algorithm}
\usepackage{float}
\algtext*{EndFor}%
\algtext*{EndProcedure}%

\usepackage{booktabs}
\usepackage[normalem]{ulem}
\useunder{\uline}{\ul}{}

\usepackage{times}
\usepackage{latexsym}

\usepackage[T1]{fontenc}
\usepackage{amsmath}
\usepackage{amssymb}
\usepackage{scalerel,xparse}
\usepackage[utf8]{inputenc}

\usepackage{cleveref}

\usepackage{microtype}
\usepackage{enumitem}
\crefformat{section}{\S#2#1#3}
\crefformat{subsection}{\S#2#1#3}
\crefformat{subsubsection}{\S#2#1#3}

\title{Mastering the ABCDs of Complex Questions: \\ Answer-Based Claim Decomposition for Fine-grained Self-Evaluation}

 \author{Nishant Balepur $\quad$ Jie Huang $\quad$ Samraj Moorjani $\quad$ Hari Sundaram $\quad$ Kevin Chen-Chuan Chang \\
 University of Illinois at Urbana-Champaign, USA \\
 \texttt{ \{balepur2,jeffhj,samrajm2,hs1,kcchang\}@illinois.edu}
}

\begin{document}
\maketitle
\begin{abstract}

When answering complex questions, large language models (LLMs) may produce answers that do not satisfy all criteria of the question. While existing self-evaluation techniques aim to detect if such answers are correct, these techniques are unable to determine which criteria of the question are satisfied by the generated answers. To address this issue, we propose \emph{answer-based claim decomposition} (\textsc{Abcd}), a prompting strategy that decomposes questions into a series of true/false claims that can be used to verify which criteria of the input question an answer satisfies. Using the decomposed \textsc{Abcd} claims, we perform fine-grained self-evaluation. Through preliminary experiments on three datasets, including a newly-collected challenge dataset \textsc{ObscureQA}, we find that GPT-3.5 has some ability to determine to what extent its answer satisfies the criteria of the input question, and can give insights into the errors and knowledge gaps of the model.\footnote{Our dataset can be found at \url{https://huggingface.co/datasets/nbalepur/ObscureQA}}

\end{abstract}

\section{Introduction}

Large language models (LLMs) have been shown to hallucinate, meaning that they generate statements that sound plausible but are untruthful \cite{alkaissi2023artificial, bang2023multitask}. Such convincing hallucinations may cause users to trust untruthful answers to questions, which could have dire consequences if LLMs are used to inform decisions \cite{10.1145/3351095.3372852, evans2021truthful}.

\begin{figure}[t]
    \centering
    \includegraphics[width=\linewidth]{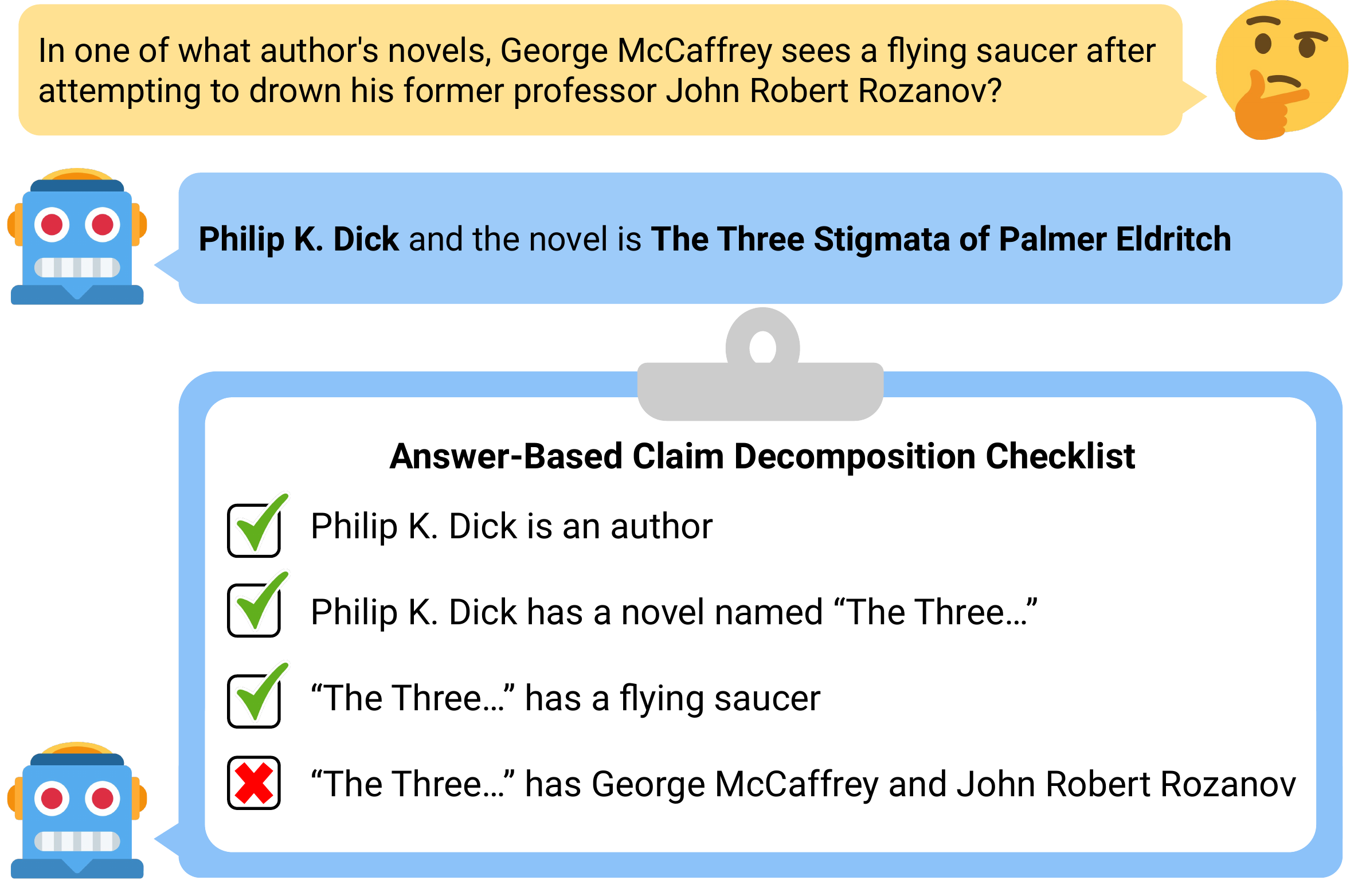}
    \vspace{-4ex}
    \caption{Using answer-based claim decomposition to verify ChatGPT's answer to an \textsc{ObscureQA} question.}
    \label{fig:intro}
    \vspace{-3.25ex}
\end{figure}

One reason LLMs may hallucinate is by failing to attend to all parts of the input question \cite{tian2019sticking, ji2023survey}. For example, in Figure~\ref{fig:intro}, the LLM incorrectly answers the provided question with Philip K. Dick. However, this answer satisfies some, but not all, of the criteria outlined in the question. While Philip K. Dick is a well-known author with a novel that contains a flying saucer (\emph{The Three Stigmata of Palmer Eldritch}), this novel does not contain the character George McCaffrey or John Robert Rozanov, as well as the described scene between these characters. Hence, it may be that the LLM mainly attends to the parts of the question relating to ``author'', ``novel'', and ``flying saucer'', while ignoring the other important parts of ``George McCaffrey'' and ``John Robert Rozanov''.

\begin{figure*}[t]
    \centering
    \includegraphics[width=\linewidth]{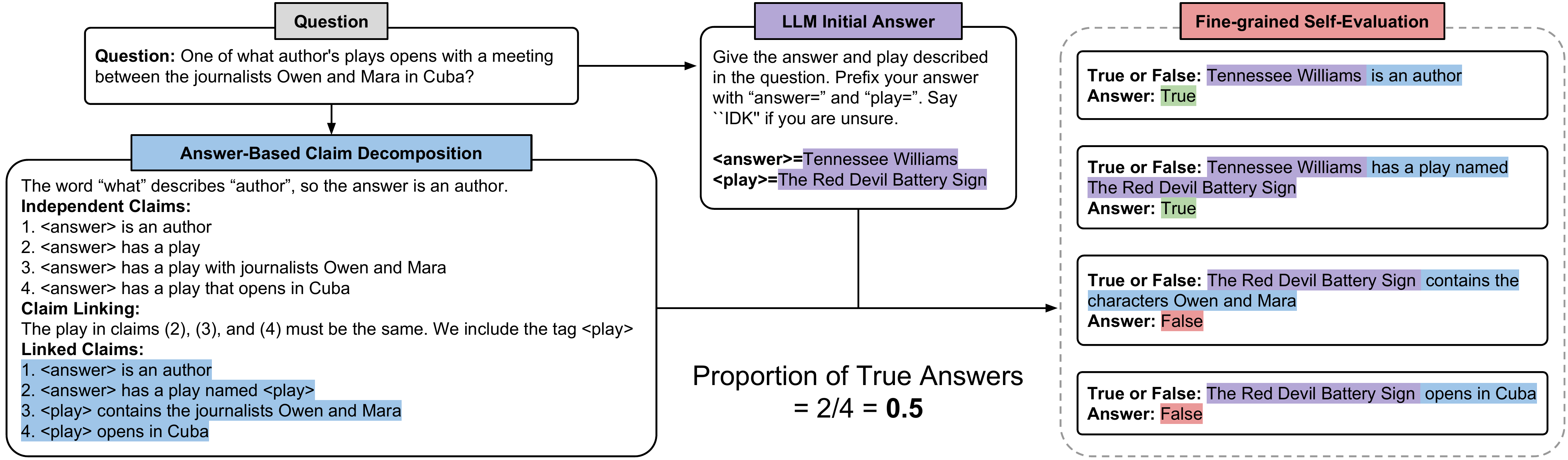}
    \vspace{-4ex}
    \caption{Overview of Fine-grained Self-Evaluation. First, given an input question, the LLM separately generates a series of claims with $\textsc{Abcd}$ and an initial answer to the question. Next, the LLM self-evaluates the generated answer with respect to the decomposed claims from $\textsc{Abcd}$. Finally, we calculate the proportion of ``true'' responses.} 
    \vspace{-2ex}
    \label{fig:model}
\end{figure*}

To detect such hallucinations from LLMs, previous works have leveraged self-evaluation, a strategy where an LLM evaluates its previously-generated answer \cite{Kadavath2022LanguageM}. However, existing self-evaluation methods typically determine if an answer is correct with respect to the entire question, without considering which criteria of the question the answer may satisfy. We intuit that directly evaluating an answer to a complex question may be difficult, but decomposing the question into smaller claims and evaluating these claims individually could be more feasible. Further, performing self-evaluation with respect to decomposed claims surrounding the question could help us analyze LLM behavior at a fine-grained level.

To manifest this idea, we first propose \emph{answer-based claim decomposition} (\textsc{\textsc{Abcd}}), a prompting strategy that generates a list of claims, comprising all criteria of the complex question, that follow from the assumption that the answer to the question is known, exhibited by the checklist in Figure~\ref{fig:intro}. Using these claims, we perform fine-grained self-evaluation, illustrated in Figure~\ref{fig:model}. After the LLM generates a response to the input question, we use the same model to self-evaluate the proportion of \textsc{Abcd} claims the LLM believes its answer satisfies. 

We test fine-grained self-evaluation with GPT-3.5 on multiple datasets, including standard trivia questions on \textsc{TriviaQA} \cite{joshi2017triviaqa}, multi-hop reasoning questions on \textsc{HotpotQA} \cite{yang-etal-2018-hotpotqa}, and obscure trivia questions on a newly-collected challenge dataset \textsc{ObscureQA}. The questions in \textsc{ObscureQA} are derived from college-level QuizBowl questions, which are written by trivia experts. Thus, these questions exploit the knowledge gaps in the training data of LLMs, prompting the LLM to produce incorrect responses that do not satisfy all criteria of the input question, as shown in Figure \ref{fig:intro}. Overall, \textsc{ObscureQA} provides a challenging testbed for fine-grained self-evaluation and other truthfulness techniques.

Our findings suggest that GPT-3.5 has some ability to verify its answers with the decomposed claims produced by \textsc{Abcd}. Specifically, there is a significant difference in the proportion of claims satisfied for incorrect and correct responses (\cref{sec:quant}), and we show how fine-grained self-evaluation can qualitatively provide deeper insights into the errors and knowledge gaps of the LLM (\cref{sec:qual}). 
Despite these findings, we observe that fine-grained self-evaluation still lacks substantial reliability. Hence, we conduct an error analysis to scrutinize its limitations and propose potential solutions (\cref{sec:error}).

\noindent Our contributions can be summarized as follows: \\
\textbf{1)} We introduce \emph{answer-based claim decomposition}, a prompting strategy to decompose questions into claims that follow from the assumption that the answer to the question is known. \\
\textbf{2)} We release \textsc{ObscureQA}, a new challenge dataset consisting of difficult trivia questions that tend to elicit untruthful responses from LLMs. \\
\textbf{3)} We use \textsc{Abcd} for fine-grained self-evaluation on three QA datasets, showcasing that GPT-3.5 has some ability to determine to what extent its answer satisfies the criteria outlined in the input question.\\

\section{Related Work}


\noindent \textbf{Problem Decomposition:} When faced with a complex problem, LLMs have shown to benefit from decomposing said problem into smaller, more manageable subproblems \cite{perez-etal-2020-unsupervised, huang2022towards}. This technique has been manifested through various prompting strategies, including least-to-most prompting \cite{zhou2023leasttomost, drozdov2023compositional}, successive prompting \cite{dua2022successive}, and decomposed prompting \cite{DBLP:journals/corr/abs-2210-02406}. \textsc{Abcd} is also a prompt-based problem decomposition technique, but rather than decomposing questions into subquestions, we decompose questions into a series of true/false claims. 

\noindent \textbf{Claim Decomposition:} \textsc{Abcd} is most similar to \citet{chen-etal-2022-generating}, who decompose political claims into a series of yes/no questions, and similarly calculate the proportion of questions with ``yes'' responses. However, using claim decomposition in question answering introduces new challenges, such as linking consistent entities in multi-hop reasoning questions (\cref{sec:abcd}). Further, \textsc{Abcd} is a prompting strategy, while \citet{chen-etal-2022-generating} fine-tune T5 to decompose claims. Another difference is that we use \textsc{Abcd} to verify LLM answers through self-evaluation, while \citet{chen-etal-2022-generating} build a retrieval system to evaluate their yes/no questions.




\noindent \textbf{LLM Self-Evaluation:} Recent work has focused on using LLMs to evaluate the veracity of their own answers. This has taken a variety of forms, including methods to quantify and calibrate uncertainty \cite{sun2022quantifying, kuhn2023semantic, cheng2023prompting} and teaching LLMs to verbalize their confidence \cite{lin2022teaching, mielke-etal-2022-reducing}. Our fine-grained self-evaluation is most similar to \citet{Kadavath2022LanguageM}, who show that LLMs can propose an answer and determine if said answer is correct. However, we evaluate answers with respect to multiple criteria of an input question, while \citet{Kadavath2022LanguageM} verify if the answer is correct with respect to the entire input question. Hence, our fine-grained self-evaluation can provide a deeper understanding of LLM behavior (\cref{sec:qual}).



\section{Method}

\subsection{Answer-Based Claim Decomposition}  \label{sec:abcd}

Given a question $q$, we aim to generate a list of claims $\mathcal{C}$ that are derived from the assumption that \texttt{<answer>} is the correct response to $q$. Each claim in $\mathcal{C}$ includes a set of tags $\mathcal{T}$ (e.g. \texttt{<answer>}, \texttt{<play>}) representing key entities that are necessary to fully answer the question $q$. 


Leveraging in-context learning \cite{brown2020language}, we propose a three-step prompt for \textsc{Abcd}, illustrated in Figure~\ref{fig:model} (left). \textbf{First}, we decompose the question into a list of independent (i.e. claims that do not rely on each other) claims that solely revolve around the answer tag $\texttt{<answer>}$. We also prompt the LLM with additional reasoning to determine the entity type of $\texttt{<answer>}$ (e.g., shown in Figure \ref{fig:model}, ``the word `what' describes an `author', so the answer is an author'').


However, these independent claims are not strict enough, as they do not consider the information that must be consistent across claims. For example, in Figure~\ref{fig:model} (left), the independent claims in (2), (3), and (4) discuss \emph{a} play, but for the answer to be correct, this play must be consistent. Hence, as a \textbf{second} step, the LLM identifies which claims discuss the same entities and include extra tags for said entities. In the \textbf{third} step, the LLM adds the extra tags to the corresponding independent claims. Steps 2 and 3 effectively allow the LLM to decompose questions that contain multi-hop reasoning. 

We use GPT-3.5 (\texttt{text-davinci-003}) to perform \textsc{Abcd}.\footnote{We found that GPT-4 (\texttt{gpt-4}) had slightly better performance when decomposing questions, but did not justify the twice as expensive cost. Although \texttt{gpt-3.5-turbo} is less expensive, the model is optimized for dialogue, and thus performed poorly when following in-context examples.} For each dataset, we manually write 5-7 examples (shown in Appendix \ref{sec:abcd_prompt}) to prompt GPT-3.5 and use a temperature of 0. 

\subsection{Fine-grained Self-Evaluation}

Prior research has demonstrated that LLMs can assess the veracity of their own answers \cite{Kadavath2022LanguageM, xie2023decomposition}. Hence, we study the ability of an LLM to determine the accuracy of \emph{its own answer} with respect to the \textsc{Abcd} claims. Motivated by \citet{chen-etal-2022-generating}, we believe that correct answers will exhibit a greater ratio of claims that the LLM determines to be true compared to incorrect ones. Further, performing self-evaluation on the decomposed claims rather than the entire question (i.e. ``Here is the question: $q$? Is the answer $a$ correct?'') can give a deeper understanding of the LLM's errors and knowledge gaps (\cref{sec:qual}).  

Given a question $q$, \textsc{Abcd} claims $\mathcal{C}$, and tags $\mathcal{T}$ found in $\mathcal{C}$, we first generate a list of answers $\mathcal{A} = (a_1, ..., a_n)$ corresponding to each tag $\mathcal{T} = (t_1, ..., t_n)$. To do so, we use a zero-shot prompt with GPT-3.5 (\texttt{gpt-3.5-turbo}), shown in Figure \ref{fig:model} (middle). We manually annotate if each initial answer is correct, as we find existing evaluation metrics insufficient \cite{si-etal-2021-whats}. 

After generating a list of answers $\mathcal{A}$ corresponding to the tags $\mathcal{T}$ found in $\mathcal{C}$, we replace each tag $t_i$ with its answer $a_i$ to obtain a list of true/false claims $\mathcal{C}_{tf}$. We ask the same GPT-3.5 model (\texttt{gpt-3.5-turbo}) to independently verify each claim $c \in \mathcal{C}_{tf}$ with the following prompt: ``\texttt{True or False: $c$}''. We decode with a temperature of 0 and write a Python script to determine whether the LLM answer was true, false, or non-responsive (i.e. ``IDK''), detailed in Appendix \ref{sec:parse_verify}.

We repeat this process for each claim $c \in \mathcal{C}_{tf}$ and calculate $\text{score}_{T}(\mathcal{C}, a)$, the average proportion of claims in $\mathcal{C}_{tf}$ the LLM determines to be true:
\begin{equation}\small
    \text{score}_{T}(\mathcal{C}, a) = \frac{1}{n-1} \sum_{i=2}^{n} \mathds{1}(\mathcal{C}_{tf}(i) = \texttt{``true''}),
\end{equation}
where $\mathds{1}$ is the indicator function and $n$ is the number of claims in $\mathcal{C}_{tf}$.\footnote{We omit the first claim ($i=1$) as it describes the entity type of the answer (e.g. ``<answer> is a person''). We found this claim to almost always be true and thus uninformative for self-evaluation.}
\begin{table}[t]
\small
\centering
\setlength{\tabcolsep}{3pt}
\begin{tabular}{@{}llccccc@{}}
\toprule
\multicolumn{1}{c|}{\textbf{Dataset}} & \multicolumn{1}{c}{\textbf{C}} & \textbf{I} & \textbf{Diff} & \multicolumn{1}{c|}{\textbf{p-val}} & \textbf{$\mathbb{P}$(C)} & \textbf{$\mathbb{P}$(I)} \\ \midrule
\multicolumn{1}{l|}{TriviaQA} & 0.887 & 0.581 & 0.306 & \multicolumn{1}{c|}{0.000} & 0.75 & 0.18 \\
\multicolumn{1}{l|}{HotpotQA (easy)} & 0.769 & 0.546 & 0.223 & \multicolumn{1}{c|}{0.002} & 0.52 & 0.34 \\
\multicolumn{1}{l|}{HotpotQA (med)} & 0.765 & 0.484 & 0.281 & \multicolumn{1}{c|}{0.000} & 0.42 & 0.34 \\
\multicolumn{1}{l|}{ObscureQA} & 0.613 & 0.494 & 0.120 & \multicolumn{1}{c|}{0.038} & 0.26 & 0.70 \\ \bottomrule
\end{tabular}
\vspace{-1.5ex}
\caption{\label{table:self_eval}Average proportion of \textsc{Abcd} claims GPT-3.5 finds true ($\text{score}_{T}(\mathcal{C}, a)$) for \textbf{C}orrect vs \textbf{I}ncorrect answers using fine-grained self-evaluation. \textbf{Diff} is the difference of the two values. \textbf{p-val}ue is for a t-test for $\text{Diff}=0$. We also show the proportion of initial answers that were correct/incorrect ($\mathbb{P}$(C)/$\mathbb{P}$(I)).}
\vspace{-2ex}
\end{table}

\section{Datasets}


We apply \textsc{Abcd} to various open-domain questions in a closed-book setting. We select 200 questions from \textsc{TriviaQA} \cite{joshi2017triviaqa} to represent traditional trivia questions, and 300 questions from \textsc{HotpotQA} \cite{yang-etal-2018-hotpotqa} (150 easy, 150 medium) to represent multi-hop reasoning questions. Further, we evaluate our technique on \textsc{ObscureQA}, a new challenge dataset with difficult trivia questions designed for evaluating truthfulness techniques such as \textsc{Abcd}. We briefly describe the collection process of \textsc{ObscureQA} below.

To curate a dataset of obscure trivia questions, we seek questions that require expert-level knowledge to answer. We find that Quizbowl questions \cite{boyd2018human, rodriguez2019quizbowl} fulfill this criterion. Quizbowl questions are comprised of a series of statements or clues, arranged from most obscure to least obscure, that describe a single answer. To obtain these questions, we web scrape QDB,\footnote{\url{https://nocard.org/}. We obtained permission from the author of the website to perform web scraping.} a popular Quizbowl question database. We scrape college-level Quizbowl questions and convert the most obscure clue of each question sequence into a single question by replacing the word ``this'' with ``what''. We collect 7278 obscure trivia questions to build \textsc{ObscureQA}. 


For the purpose of a preliminary study of \textsc{Abcd}, we select 200 questions from the validation set of \textsc{ObscureQA}, but hope future research can leverage \textsc{ObscureQA} at a larger scale to study LLM truthfulness. For each question $q$ in our datasets, we perform $\textsc{Abcd}$ to generate a list of claims $\mathcal{C}$, and fix these claims for the rest of our experiments.

\begin{table*}[]
\small
\centering
\resizebox{\textwidth}{!}{%
\begin{tabular}{@{}c|l|l@{}}
\toprule
\textbf{Error} & \multicolumn{1}{c|}{\textbf{Example}} & \multicolumn{1}{c}{\textbf{Correction}} \\ \midrule
\multicolumn{1}{c|}{Self-Consistency} & \begin{tabular}[c]{@{}l@{}}\textbf{Claim:} Robert Downey Jr. was born on \textbf{\textcolor{red}{December 4, 1964}}\\ \textbf{Response:} True\end{tabular} & \multicolumn{1}{l}{\begin{tabular}[c]{@{}l@{}}\textbf{Question:} When was Robert Downey Jr. born?\\ \textbf{Response:} Robert Downey Jr. was born on \textbf{\textcolor{blue}{April 4, 1965.}}\end{tabular}} \\ \midrule
\multicolumn{1}{c|}{Tense} & \begin{tabular}[c]{@{}l@{}}\textbf{Claim:} Vienna \textbf{\textcolor{red}{has}} one bishop named Melchior Klesl\\ \textbf{Response:} False. Vienna currently has a cardinal\\ named Christoph Schuborn...\end{tabular} & \multicolumn{1}{l}{\begin{tabular}[c]{@{}l@{}}\textbf{Claim:} Vienna \textbf{\textcolor{blue}{had}} one bishop named Melchior Klesl\\ \textbf{Response:} True\end{tabular}} \\ \midrule
\multicolumn{1}{c|}{Ambiguity} & \begin{tabular}[c]{@{}l@{}}\textbf{Question:} One tower in what country...\\ \textbf{Claim:} United Arab Emirates has \textbf{\textcolor{red}{one}} tower named Burj Al Arab\\ \textbf{Response:} False. United Arab Emirates has multiple towers...\end{tabular} & \multicolumn{1}{l}{\begin{tabular}[c]{@{}l@{}}\textbf{Claim:} United Arab Emirates has \textbf{\textcolor{blue}{a}} tower named Burj Al Arab\\ \textbf{Response:} True\end{tabular}} \\ \midrule
\multicolumn{1}{c|}{Reasoning} & \begin{tabular}[c]{@{}l@{}}\textbf{Claim:} Bishop Jordan \textbf{\textcolor{red}{was responsible}} for \\ converting Poland to Christianity\\ \textbf{Response:} False.  Bishop Jordan was not responsible...\end{tabular} & \multicolumn{1}{l}{\begin{tabular}[c]{@{}l@{}}\textbf{Evidence 1:} The \textbf{\textcolor{blue}{impetus}} to the process [Christianization of Poland] \\ was ... the \textbf{\textcolor{blue}{personal baptism of Mieszko I}}\\ \textbf{Evidence 2:} He [\textbf{\textcolor{blue}{Bishop Jordan}}] arrived in the territory of Lechina\\ (the future Poland) ... \textbf{\textcolor{blue}{to baptise Mieszko I}}\end{tabular}} \\ \midrule
\multicolumn{1}{c|}{False Assumptions} & \begin{tabular}[c]{@{}l@{}}\textbf{Claim:} The Shipping News's character Quoyle reveals\\ his sheltered white lifestyle \textbf{\textcolor{red}{by declaring}} that there were\\ "no immigrants" in the United States during his childhood.\\ \textbf{Response:} True. Quoyle's statement reveals his sheltered \\white lifestyle because...\end{tabular} & \multicolumn{1}{l}{\begin{tabular}[c]{@{}l@{}}\textbf{Claim:} The Shipping News's character Quoyle \textbf{\textcolor{blue}{declares}} that \\ there were "no immigrants" in the United States during his childhood.\\ \textbf{Response:} False\end{tabular}} \\ \bottomrule
\end{tabular}}
\vspace{-2ex}
\caption{\label{table:error} Examples of GPT-3.5 errors during answer-based claim decomposition and fine-grained self-evaluation. Text in \textbf{\textcolor{red}{red}} indicates the error, while text in \textbf{\textcolor{blue}{blue}} indicates the prompt change or solution to overcome the error. ``\textbf{Response:}'' indicates the GPT-3.5 response using a temperature of 0. ``\textbf{Evidence:}'' comes from Wikipedia. All displayed claims were produced by answer-based claim decomposition technique on our three datasets.}
\vspace{-2ex}
\end{table*}
\begin{figure}
    \centering
    \fbox{\includegraphics[width=0.9\linewidth]{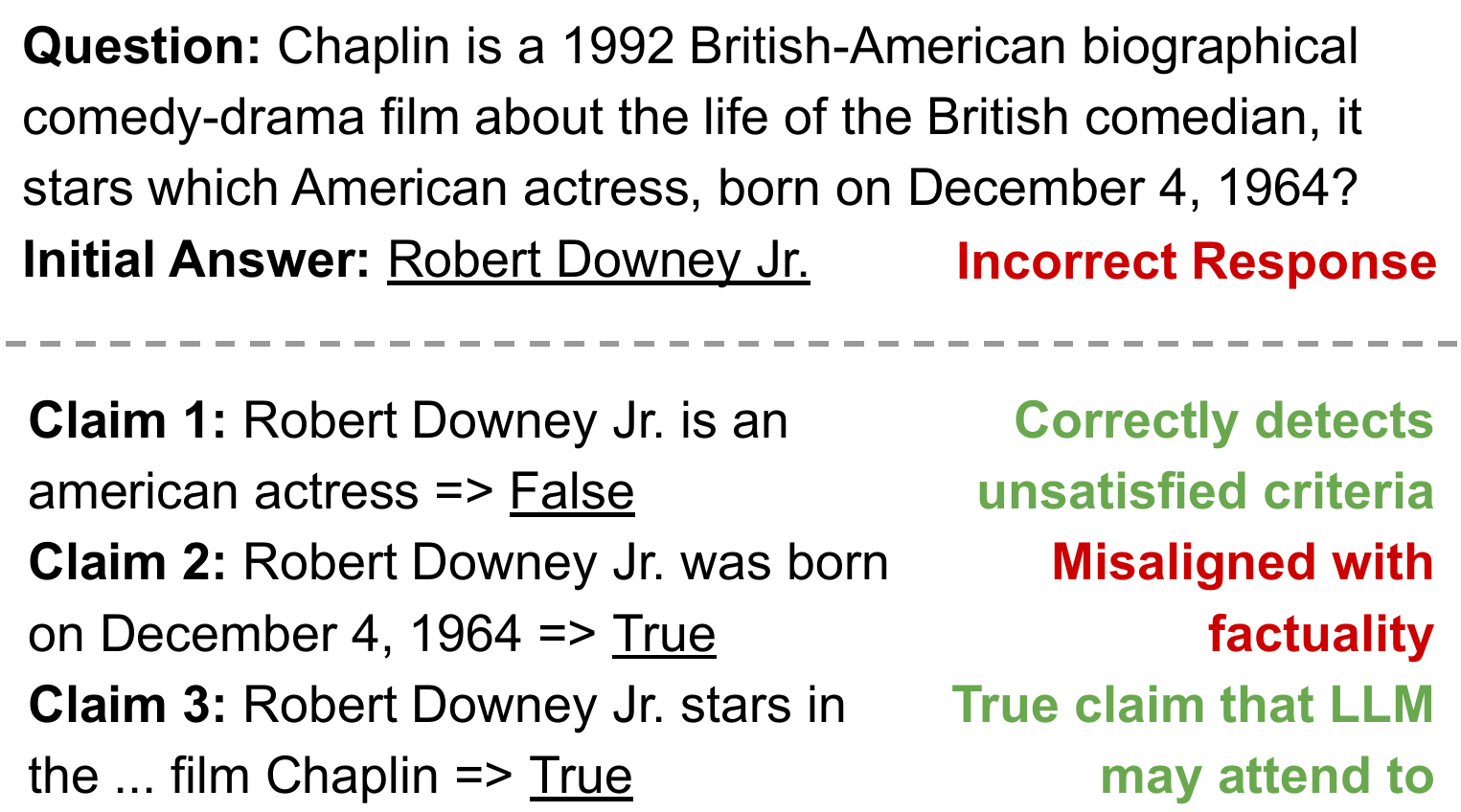}}
    \vspace{-1ex}
    \caption{\label{fig:qual}Qualitative analysis of fine-grained self-evaluation on a question from \textsc{TriviaQA}.}
    \vspace{-2ex}
\end{figure}

\section{Results}

\subsection{Quantitative Results} \label{sec:quant}

In Table \ref{table:self_eval}, on all datasets, including the challenge dataset \textsc{ObscureQA}, we observe a significant difference in the average $\text{score}_{T}(\mathcal{C}, a)$ calculated when $a$ is correct versus incorrect. This result suggests that GPT-3.5 has some ability to determine to what extent its answer satisfies all criteria outlined in the complex input question. Finally, we note that \textsc{ObscureQA} is truly a challenging dataset, eliciting the most untruthful responses from GPT-3.5 ($\mathbb{P}(\text{I})=0.7$).



\subsection{Qualitative Analysis} \label{sec:qual}

In Figure \ref{fig:qual}, we demonstrate how fine-grained self-evaluation can lead to a deeper understanding of LLM errors. Although Robert Downey Jr is incorrect, GPT 3.5 clarifies that he is not an American actress, and thus does not fulfill all criteria of the input question. Further, we see that the model likely arrived at Robert Downey Jr because he starred in the film Chaplin, and the model is misaligned with the true birthdate of the actor. We provide more examples in Appendix~\ref{sec:self_eval_comp} and show how fine-grained self-evaluation can be more informative than evaluating with respect to the entire question.

\subsection{Error Analysis} \label{sec:error}

In section \cref{sec:quant}, we found that GPT-3.5 has some ability to evaluate its own answer with respect to the \textsc{Abcd} claims. To fully investigate the reliability of our approach, we conduct an error analysis. In Table \ref{table:error}, we display some of the most prevalent categories of errors along with potential corrections. We briefly describe each error type below:

\subsubsection{Self-Consistency}

We find that GPT-3.5 exhibits inconsistencies in its beliefs when determining the truthfulness of claims, which has also been noted in prior work \cite{hase-etal-2023-methods}. For example, GPT-3.5 incorrectly claims that Robert Downey Jr. was born on December 4, 1964, but when separately asked to provide the actor's birth date, the LLM correctly states it is April 4, 1965. To overcome this issue, we suggest designing a more robust claim verification technique, such as an ensemble of prompts to measure consistency under paraphrase \cite{elazar-etal-2021-measuring, de-cao-etal-2021-editing}, or incorporating fact verification metrics that involve question generation/answering \cite{wang-etal-2020-asking}.

\subsubsection{Tense}

In some cases, we find that GPT-3.5 fails to preserve the tense of the question during claim decomposition, even when appropriate examples are included in the \textsc{Abcd} prompt. We find that preserving verb tense is essential during claim verification, as rewording a claim from the present to the past tense can change the implication of the claim. For example, the claim ``Vienna has one bishop named Melchior Klesl'' is false because Melchior Klesl is not the current bishop of Vienna, but changing ``has'' to ``had'' will make the claim true. Although LLMs have been shown to understand linguistic phenomena such as tense to some extent \cite{zhang-etal-2022-probing}, we believe it would be beneficial to leverage chain-of-thought style prompting \cite{wei2022chain} to help the LLM determine the necessary verb tense of each claim. 

\subsubsection{Ambiguity}

When decomposing claims, we find that GPT-3.5 may adhere to the wording of the question too closely, resulting in ambiguous claims that are difficult to evaluate. For example, the claim ``UAE has one tower named Burj Al Arab'' could be interpreted as ``\emph{the UAE has \textbf{exactly one tower}, and the name of this tower is Burj Al Arab}'' or ``\emph{the UAE has \textbf{a} tower named Burj Al Arab}''. The latter is how we want the claim to be interpreted, but we find GPT-3.5 interprets the claim as the former. Studying ways to help LLMs model ambiguity is essential to overcome this weakness \cite{liu2023we}.

\subsubsection{Reasoning}

If a claim requires complex reasoning, we find that GPT-3.5 fails to assess the claim accurately. For example, GPT-3.5 states that the claim ``Bishop Jordan was responsible for converting Poland to Christianity'' is false, but evidence\footnote{\url{https://en.wikipedia.org/wiki/Christianization_of_Poland}} suggests that (1) the conversion of Poland to Christianity was sparked by the baptism of Mieszko I; and (2) Bishop Jordan baptized Mieszko I. Performing reasoning over this evidence could lead the LLM to determine that the claim is true. While obtaining and leveraging such evidence from the web is feasible \cite{DBLP:journals/corr/abs-2112-09332}, it would defeat the purpose of self-evaluation. Hence, we believe the best way to address this problem is by equipping LLMs with advanced reasoning techniques when verifying claims \cite{huang2022towards}.

\subsubsection{False Assumptions}

We find that some of the decomposed claims contain false assumptions, which are challenging for LLMs to address \cite{Kim2022QA2QA}. For example, the original claim in Table \ref{table:error} has the false assumption that Quoyle’s quoted statement exists in the novel, causing GPT-3.5 to agree that the statement ``reveals his sheltered white lifestyle.'' However, we find that if we decompose the claim further to question if the statement even exists in the novel, GPT-3.5 can accurately determine that the statement does not exist. Hence, we believe the issue of false assumptions could be overcome by performing $\textsc{Abcd}$ recursively until each claim is determined to be fully decomposed.

\subsection{Ground Truth Comparison}

When an LLM generates an incorrect response, it is beneficial to know if the LLM \emph{could have} generated the correct response based on its internal knowledge \cite{cohen2023crawling}. Hence, for each predicted incorrect response $a_{pred}$, we compare $\text{score}_{T}(\mathcal{C}, a_{pred})$ and $\text{score}_{T}(\mathcal{C}, a_{gt})$, where $a_{gt}$ is the ground truth answer. If $\text{score}_{T}(\mathcal{C}, a_{pred}) < \text{score}_{T}(\mathcal{C}, a_{gt})$, it indicates that the LLM has the knowledge needed to answer the question correctly. 

In Table \ref{table:gt}, we find that there are few instances where GPT-3.5 determines the ground truth answer satisfies more \textsc{Abcd} claims compared to its incorrect generated answer. Hence, in our experiments, although GPT-3.5 can identify when its answer is incorrect to an extent, there is no strong evidence that the model can answer the question correctly.

\begin{table}[]
\small
\centering
\setlength{\tabcolsep}{3pt}
\begin{tabular}{@{}l|c|c|c@{}}
\toprule
\multicolumn{1}{c|}{\textbf{Dataset}} & \textbf{\# GT > Pred} & \textbf{\# GT = Pred} & \textbf{\# GT < Pred} \\ \midrule
TriviaQA & 6 & 21 & 8 \\
HotpotQA (easy) & 8 & 35 & 8 \\ \bottomrule
\end{tabular}
\vspace{-1.5ex}
\caption{\label{table:gt}Number of times when $\text{score}_{T}(\mathcal{C}, a_{gt})$ ($\textbf{GT}$) is greater than/equal to/less than $\text{score}_{T}(\mathcal{C}, a_{pred})$ ($\textbf{Pred}$).}
\vspace{-2.5ex}
\end{table}
\section{Conclusion}

We introduce \emph{answer-based claim decomposition}, which aims to decompose a question into a series of true/false claims. Through experiments on three datasets with GPT-3.5, including a new challenge dataset \textsc{ObscureQA}, we show how our technique can be used to perform fine-grained self-evaluation. We find that there is a significant difference in the proportion of claims satisfied for incorrect and correct responses, but there is no indication that GPT-3.5 believes that the gold answer satisfies more \textsc{Abcd} claims than its incorrect answers. Finally, to investigate the reliability of our approach, we conduct an error analysis and based on our findings, suggest remedies to overcome these errors.



\section{Limitations}

In our preliminary experiments, we apply answer-based claim decomposition to factual trivia questions where answers are entities spanning a few words. However, we did not examine if our technique would be effective on other types of QA datasets, such as \textsc{TruthfulQA} \cite{lin-etal-2022-truthfulqa}, which exploits imitative falsehoods and contains longer desired responses, or \textsc{BoolQA} \cite{clark-etal-2019-boolq}, which has ``yes'' or ``no'' as the only possible answers. 

Further, due to financial constrains, we test \textsc{Abcd} and fine-grained self-evaluation through preliminary experiments on a subset of data from our three datasets. However, given that our results were statistically significant (\cref{sec:quant}), we believe that the number of questions selected were sufficient for our study. In addition, since we only examined a subset of questions from our newly-collected dataset \textsc{ObscureQA}, this opens up future research directions leveraging our dataset. We believe that \textsc{ObscureQA} could be used to evaluate LLMs on a variety of facets, including benchmarking the academic knowledge of state-of-the-art LLMs, and given that this dataset frequently elicits untruthful responses, studying confidence and uncertainty calibration techniques. 


\section{Ethics Statement}

Combating hallucinations is a key step to ensuring LLMs are aligned with factuality and truthfulness. In this work, we showcase how answer-based claim decomposition can be used to determine to what extent an LLM output satisfies the criteria of the input question. However, since our strategy is a self-evaluation technique that must be used after an answer is generated, it would also be beneficial to explore if $\textsc{Abcd}$ can be used to improve factuality and truthfulness during the first pass of generation.

Given that GPT-3.5 recognizes that its answers do not satisfy all claims, one promising direction is to optimize question answering on satisfying the \textsc{Abcd} claims themselves. If all claims cannot be satisfied with a single answer, the LLM could respond with uncertainty (e.g. ``IDK''). Further, we do not currently use any weighting to determine the importance of claims. Some claims may be more important than others when evaluating an answer, so ranking claims by relevance, specificity, or difficulty could help the LLM efficiently reason towards a factual and truthful answer. Overall, we believe that \textsc{Abcd} and \textsc{ObscureQA} are promising tools for developing truthful and honest LLMs.

\bibliography{anthology,custom}
\bibliographystyle{acl_natbib}

\clearpage

\appendix

\begin{figure*}
    \centering
    \fbox{\includegraphics[width=0.7\linewidth]{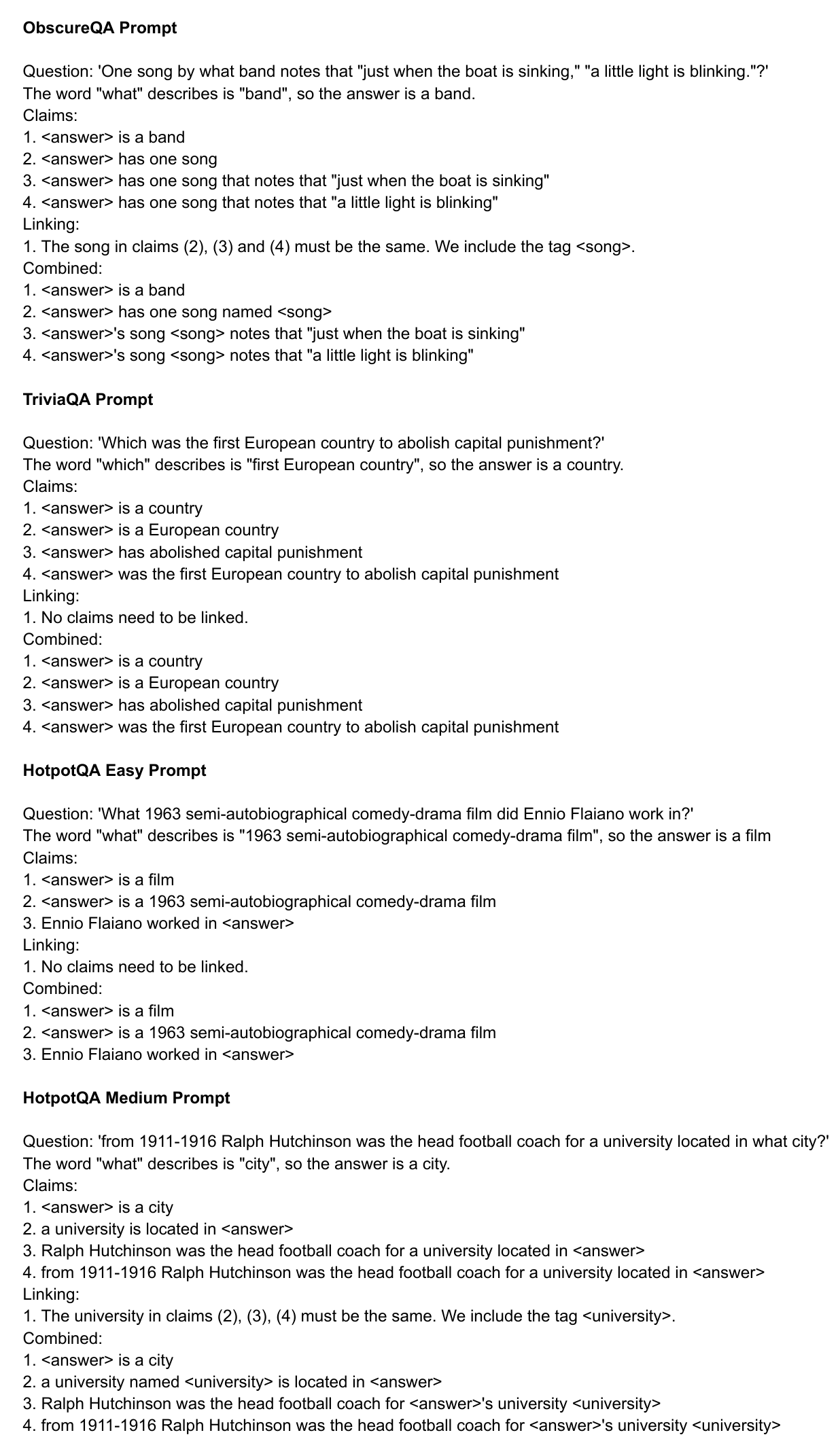}}
    \caption{In-context learning examples used to decompose questions into claims with $\textsc{Abcd}$ on our QA datasets. These claims are used in fine-grained self-evaluation.}
    \label{fig:ours}
\end{figure*}


\begin{figure*}
    \centering
    \fbox{\includegraphics[width=0.7\linewidth]{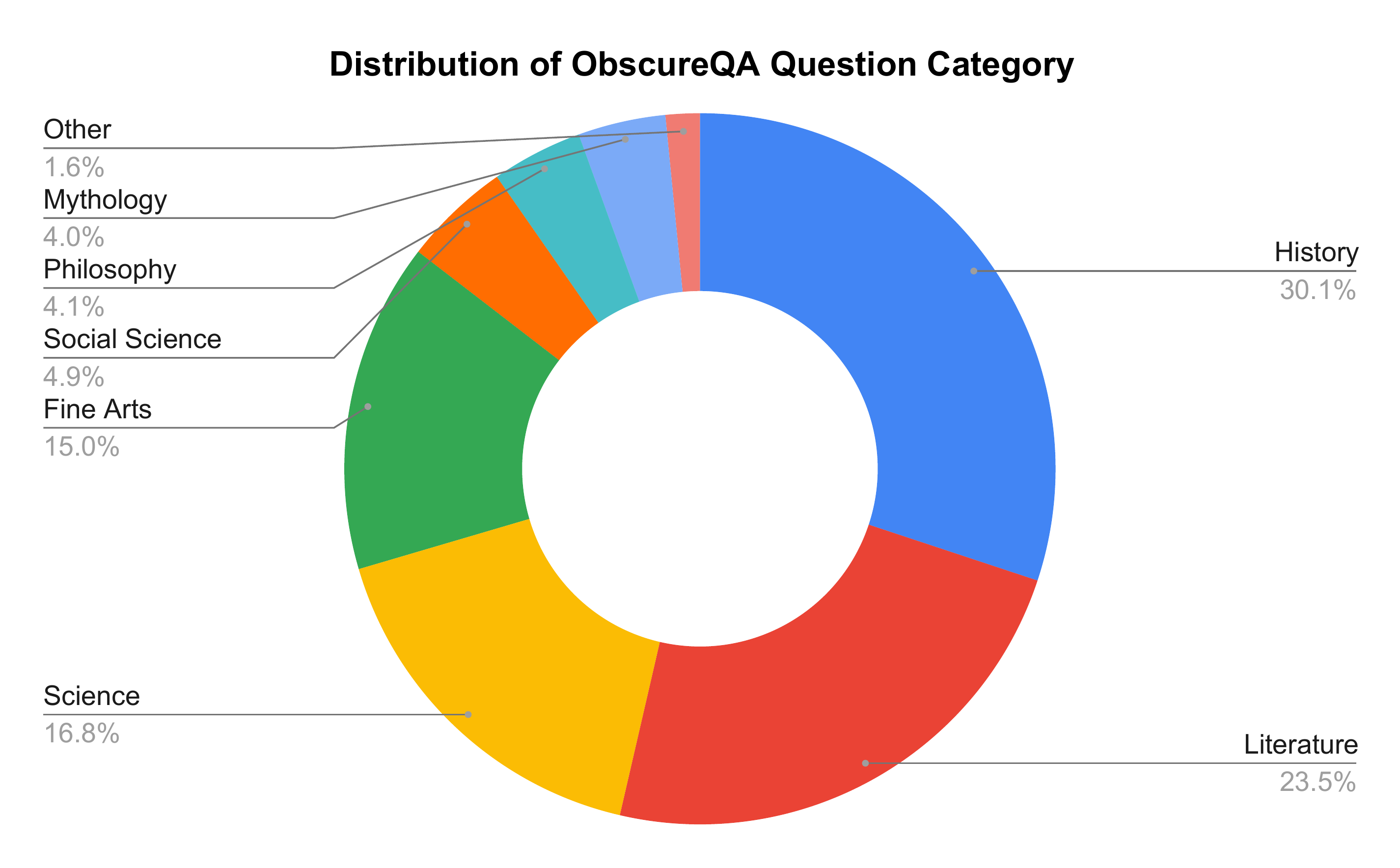}}
    \caption{Distribution of question category on \textsc{ObscureQA}. Other encompasses geography, religion, trash (pop culture), and current events.}
    \label{fig:pie}
\end{figure*}

\begin{table*}[]
\small
\centering
\begin{tabular}{@{}llllll@{}}
\toprule
\textbf{\# Train}        & \textbf{\# Valid}       & \textbf{\# Test}         & \textbf{\# Unique Categories} & \textbf{Avg. Question Length (words)} & \textbf{Avg. Answer Length (words)} \\ \midrule
\multicolumn{1}{c}{5094} & \multicolumn{1}{c}{730} & \multicolumn{1}{c}{1454} & \multicolumn{1}{c}{11}        & \multicolumn{1}{c}{28.479}            & \multicolumn{1}{c}{2.471}           \\ \bottomrule
\end{tabular}
\caption{Summary statistics of \textsc{ObscureQA}. The number of words are calculated using the nltk word tokenizer.}
\label{table:summary}
\end{table*}

\begin{figure*}
    \centering
    \fbox{\includegraphics[width=\linewidth]{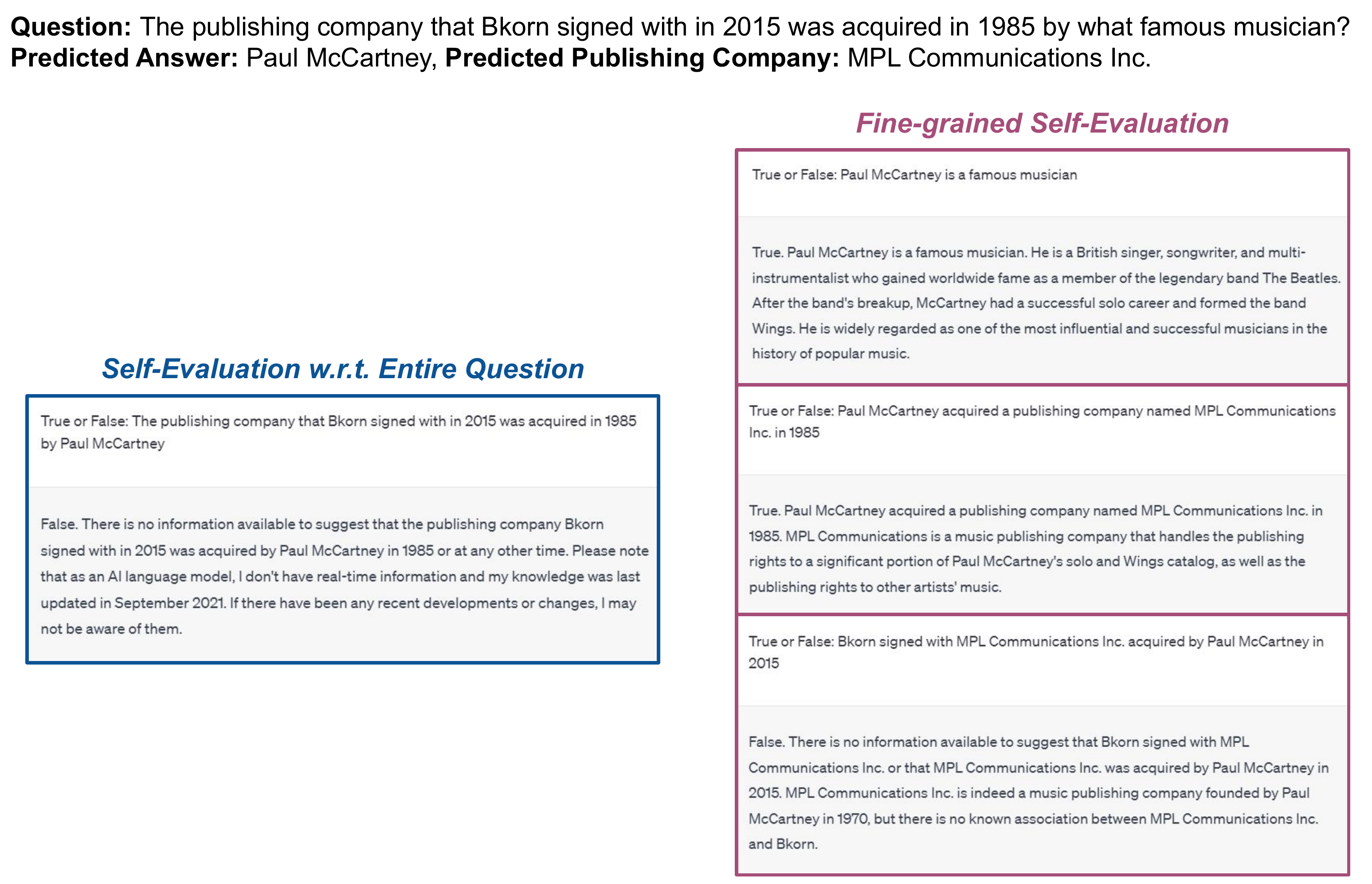}}
    \caption{Qualitative comparison of self-evaluation with respect to the entire question and fine-grained self-evaluation on a question from \textsc{HotpotQA} medium using the ChatGPT web interface. From a human's perspective, the former is uninformative, since ChatGPT essentially states the claim is false and then restates the claim with a negation (i.e. ``There is no information to suggest that [claim]''). However, in fine-grained self-evaluation, we find that ChatGPT believes the claim is false because the LLM believes there is no association between Bkorn and MPL Communications Inc. Further, we can uncover a misaligned fact in the second claim of fine-grained self-evaluation, since MPL Communications Inc. was established around 1968. }
    \label{fig:hotpotqa_qual}
\end{figure*}

\begin{figure*}
    \centering
    \fbox{\includegraphics[width=\linewidth]{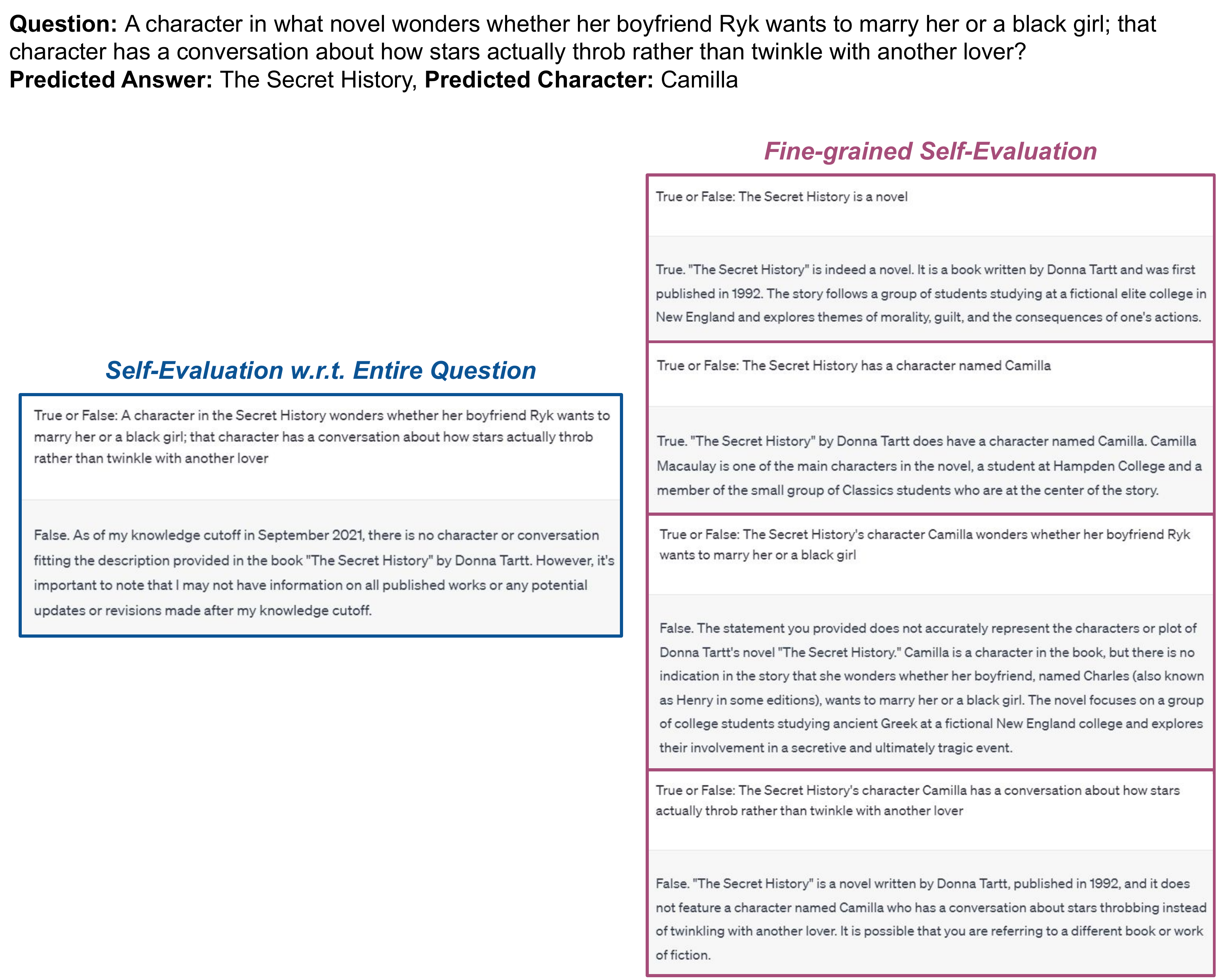}}
    \caption{Qualitative comparison of self-evaluation with respect to the entire question and fine-grained self-evaluation on a question from \textsc{ObscureQA} using the ChatGPT web interface. From a human's perspective, the former is uninformative, since ChatGPT just states that there is no character or conversation that fits the description in the question. However, in fine-grained self-evaluation, we find that ChatGPT can clarify that the character Ryk does not exist in the novel (since the boyfriend's name is Charles/Henry), and the mentioned conversation about stars in the question does not exist in the novel.}
    \label{fig:obscure_qual}
\end{figure*}

\section{Implementation}

\subsection{\textsc{Abcd} Prompting} \label{sec:abcd_prompt}

In Figure \ref{fig:ours}, we display a single in-context learning example that was used to perform answer-based claim decomposition on each of our datasets.



\subsection{Claim Verification Parsing} \label{sec:parse_verify}

When performing self-evaluation, we use the prompt ``\texttt{True or False:} $c$'', where $c$ is the claim of interest. To determine whether the response of GPT-3.5 is true, false, or non-responsive, we first determine whether the words ``true'' or ``false'' are a substring of the lowercase version of the response. If ``true'' is present, we map the response to ``true'', and similarly for ``false''. If the words ``true'' or ``false'' are not present in the entire output (with a maximum length of 64 tokens), we determine the response to be a non-response (i.e. ``IDK''), since nearly all of these responses started with the phrase ``As an AI language model...''. 

We also found an interesting behavior where for certain claims, GPT-3.5 would state that the claim was false but then restate the claim as if it were true. For example, with the prompt ``True or False: LaFayette is located in Onondaga County, New York, United States'', we found decoding with a temperature of 0 resulted in the output ``False. LaFayette is located in Onondaga County, New York, United States''. In these rare scenarios, we map the response to ``true''.

\section{Datasets}

\subsection{\textsc{TriviaQA} and \textsc{HotpotQA}}

For \textsc{TriviaQA}, we randomly select 200 question/answer pairs from the training set. For both \textsc{HotpotQA} easy and \textsc{HotpotQA} medium, we randomly selected 150 question/answer pairs from the training sets. We only selected questions that were labeled as ``bridge'' questions, and ommitted the ``comparison'' questions. When performing few-shot prompting on these datasets (\textsc{Abcd} and converting question/answer pairs into statements), we selected in-context examples that were not present in the subset of data we used for evaluation.

\subsection{\textsc{ObscureQA} Dataset Description}

When creating the \textsc{ObscureQA} dataset, we collect the question, answer, category, and subcategory. To collect this data, we web scrape the QDB website using Selenium with the permission of the author of the website. We clean the text in the question and answer with unidecode and remove text between parentheses, square brackets, and angle brackets. We also omit questions that begin with phrases similar to ``Note to moderator'', as these cannot be converted to trivia questions. After this cleaning process, we create a 70/10/20 train/validation/test split. In Table \ref{table:summary}, we display summary statistics of the \textsc{ObscureQA} dataset and in Figure \ref{fig:pie}, we display the distribution of questions by category type.

\section{Why Human Evaluation is Necessary}

The two metrics we considered for automatically evaluating the initial answer generation of GPT-3.5 were exact match and accuracy. However, these methods were insufficient in our closed-book question answering setting. For example, given the question ``What film edited by Zene Baker was co-directed by Evan Goldberg?'' on \textsc{HotpotQA} medium, the gold answer is ``The Interview.'' However, GPT-3.5 proposed an alternative answer of ``This is the End'', which is another film that was edited by Zene Barker and co-directed by Evan Goldberg. Given that both exact match and accuracy would not be able to detect this answer as correct, we decided to manually annotate the answers. 

\section{Self Evaluation Comparison} \label{sec:self_eval_comp}

\label{sec:appendix}

In Figures \ref{fig:hotpotqa_qual} and \ref{fig:obscure_qual}, we display examples of how fine-grained self-evaluation can be used to understand LLM behavior, and how this technique can be more informative compared to evaluating with respect to the entire question. Hence, we hope that future work can leverage $\textsc{Abcd}$ and fine-grained self-evaluation in a user study to see how these responses affect the user's perception of the LLM's initial response.

\end{document}